\documentclass[12pt,hidelinks]{article}
	\usepackage[explicit]{titlesec}
	\usepackage{titletoc}
	\usepackage{tocloft}
	\usepackage{charter}
	\usepackage[many]{tcolorbox}
	\usepackage{amsmath}
	\usepackage{graphicx}
	\usepackage{xcolor}
	\usepackage{tikz,lipsum,lmodern}
	\usetikzlibrary{calc}
	\usepackage[english]{babel}
	\usepackage{fancyhdr}
	\usepackage{mathrsfs}
	\usepackage{empheq}
	\usepackage{lmodern}%
        \usepackage{gensymb}
	\usepackage{scrextend}%
	\usepackage{wrapfig}
	\usepackage{fancyref}
	\usepackage{hyperref}
	\usepackage{cleveref}
	\usepackage{listings}
	\usepackage{varwidth}
	\usepackage{longfbox}
	\usepackage{geometry}
	\usepackage{marginnote}
	\tcbuselibrary{theorems}
	\tcbuselibrary{breakable, skins}
	\tcbuselibrary{listings, documentation}
\graphicspath{{./images/}}
\newcommand*\vtick{\textsc{\char13}}

\addto{\captionsenglish}{}
\renewcommand{\thesection}{\arabic{section}}
\newcommand\sectionnumfont{
	\fontsize{114}{39}\color{myblueii}\selectfont}
\newcommand\sectionnamefont{
	\normalfont\color{white}\scshape\small\bfseries }
\definecolor{mordantred19}{rgb}{0.68, 0.05, 0.0}
\definecolor{st.patrick\'sblue}{rgb}{0.14, 0.16, 0.48}
\definecolor{teal}{rgb}{0.0, 0.5, 0.5}
\definecolor{beaublue}{rgb}{0.74, 0.83, 0.9}
\definecolor{mybluei}{RGB}{0,173,239}
\definecolor{myblueii}{RGB}{63,200,244}
\definecolor{myblueiii}{RGB}{199,234,253}
\definecolor{blond}{rgb}{0.98, 0.94, 0.75}
\definecolor{cream}{rgb}{1.0, 0.99, 0.82}
\definecolor{emerald}{rgb}{0.31, 0.78, 0.47}
\definecolor{darkspringgreen}{rgb}{0.09, 0.45, 0.27}
\definecolor{ghostwhite}{rgb}{0.97, 0.97, 1.0}
\definecolor{splashedwhite}{rgb}{1.0, 0.99, 1.0}
\definecolor{whitesmoke}{rgb}{0.96, 0.96, 0.96}
\definecolor{lightgray}{rgb}{0.92, 0.92, 0.92}
\definecolor{floralwhite}{rgb}{1.0, 0.98, 0.94}
\definecolor{mygray}{gray}{0.6}
\titleformat{\section}
{\normalfont\huge\filleft}
{}
{20pt}
{\begin{tikzpicture}[remember picture,overlay]
	\fill[myblueiii] 
	(current page.north west) rectangle ([yshift=-6cm]current page.north east);   
\node[
	fill=mybluei,
	text width=2\paperwidth,
	rounded corners=3cm,
	text depth=6cm,
	anchor=center,
	inner sep=0pt] at (current page.north east) (parttop)
	{\thepart};%
\node[
	anchor=south east,
	inner sep=0pt,
	outer sep=0pt] (partnum) at ([xshift=-20pt]parttop.south) 
	{\sectionnumfont\thesection};
\node[
	anchor=south,
	inner sep=0pt] (partname) at ([yshift=2pt]partnum.south)   
	{\sectionnamefont SECTION};
\node[
	anchor=north east,
	align=right,
	inner xsep=0pt] at ([yshift=-0.5cm]partname.east|-partnum.south) 
	{\parbox{.7\textwidth}{\raggedleft#1}};
\end{tikzpicture}%
}
\hypersetup{
	colorlinks,
	linkcolor={red!50!black},
	citecolor={blue!50!black},
	urlcolor={blue!80!black}
}
\tcbset{
	defstyle/.style={
		fonttitle=\bfseries\upshape, 
		fontupper=\slshape,
		arc=0mm, 
		beamer,
		colback=blue!5!white,
		colframe=blue!75!black},
	theostyle/.style={
		fonttitle=\bfseries\upshape, 
		fontupper=\slshape,
		colback=red!10!white,
		colframe=red!75!black},
	visualstyle/.style={
		height=6.5cm,
		breakable,
		enhanced,
		leftrule=0pt,
		rightrule=0pt,
		bottomrule=0pt,
		outer arc=0pt,
		arc=0pt,
		colframe=mordantred19,
		colback=lightgray,
		attach boxed title to top left,
		boxed title style={
			colback=mordantred19,
			outer arc=0pt,
			arc=0pt,
			top=3pt,
			bottom=3pt,
		},
		fonttitle=\sffamily,},
	discussionstyle/.style={
		height=6.5cm,
		breakable,
		enhanced,
		rightrule=0pt,
		toprule=0pt,
		outer arc=0pt,
		arc=0pt,
		colframe=mordantred19,
		colback=lightgray,
		attach boxed title to top left,
		boxed title style={
			colback=mordantred19,
			outer arc=0pt,
			arc=0pt,
			top=3pt,
			bottom=3pt,
		},
		fonttitle=\sffamily},
	mystyle/.style={
		height=6.5cm,
		breakable,
		enhanced,
		rightrule=0pt,
		leftrule=0pt,
		bottomrule=0pt,
		outer arc=0pt,
		arc=0pt,
		colframe=mordantred19,
		colback=lightgray,
		attach boxed title to top left,
		boxed title style={
			colback=mordantred19,
			outer arc=0pt,
			arc=0pt,
			top=3pt,
			bottom=3pt,
		},
		fonttitle=\sffamily},
	aastyle/.style={
			height=3.5cm,
			enhanced,
			colframe=teal,
			colback=lightgray,
			colbacktitle=floralwhite,
			fonttitle=\bfseries,
			coltitle=black,
		attach boxed title to top center={
	  		yshift=-0.25mm-\tcboxedtitleheight/2,
	   		yshifttext=2mm-\tcboxedtitleheight/2}, 
		boxed title style={boxrule=0.5mm,
			frame code={ \path[tcb fill frame] ([xshift=-4mm]frame.west)
				-- (frame.north west) -- (frame.north east) -- ([xshift=4mm]frame.east)
				-- (frame.south east) -- (frame.south west) -- cycle; },
			interior code={ 
				\path[tcb fill interior] ([xshift=-2mm]interior.west)
				-- (interior.north west) -- (interior.north east)
				-- ([xshift=2mm]interior.east) -- (interior.south east) -- (interior.south west)
				-- cycle;} }
				},
	examstyle/.style={
		height=9.5cm,
		breakable,
		enhanced,
		rightrule=0pt,
		leftrule=0pt,
		bottomrule=0pt,
		outer arc=0pt,
		arc=0pt,
		colframe=mordantred19,
		colback=lightgray,
		attach boxed title to top left,
		boxed title style={
			colback=mordantred19,
			outer arc=0pt,
			arc=0pt,
			top=3pt,
			bottom=3pt,
		},
		fonttitle=\sffamily},
	doc head command={
		interior style={
			fill,
			left color=yellow!20!white, 
			right color=white}},
	doc head environment={
		boxsep=4pt,
		arc=2pt,
		colback=yellow!30!white,
		},
	doclang/environment content=text
}
\newtcolorbox[auto counter,number within=section]{example}[1][]{
	mystyle,
	title=Example~\thetcbcounter,
	overlay unbroken and first={
		\path
		let
		\p1=(title.north east),
		\p2=(frame.north east)
		in
		node[anchor=
			west,
			font=\sffamily,
			color=st.patrick\'sblue,
			text width=\x2-\x1] 
		at (title.east) {#1};
	}
}
\newtcolorbox[auto counter,number within=section]{longexample}[1][]{
	examstyle,
	title=Example~\thetcbcounter,
	overlay unbroken and first={
		\path
		let
		\p1=(title.north east),
		\p2=(frame.north east)
		in
		node[anchor=
		west,
		font=\sffamily,
		color=st.patrick\'sblue,
		text width=\x2-\x1] 
		at (title.east) {#1};
	}
}
\newtcolorbox[auto counter,number within=section]{example2}[1][]{
	aastyle,
	title=Example~\thetcbcounter,{}
}
\newtcolorbox[auto counter,number within=section]{discussion}[1][]{
	discussionstyle,
	title=Discussion~\thetcbcounter,
	overlay unbroken and first={
		\path
		let
		\p1=(title.north east),
		\p2=(frame.north east)
		in
		node[anchor=
		west,
		font=\sffamily,
		color=st.patrick\'sblue,
		text width=\x2-\x1] 
		at (title.east) {#1};
	}
}
\newtcolorbox[auto counter,number within=section]{visualization}[1][]{
	visualstyle,
	title=Visualization~\thetcbcounter,
	overlay unbroken and first={
		\path
		let
		\p1=(title.north east),
		\p2=(frame.north east)
		in
		node[anchor=
		west,
		font=\sffamily,
		color=st.patrick\'sblue,
		text width=\x2-\x1] 
		at (title.east) {#1};
	}
}
\newtcbtheorem[number within=subsection,crefname={definition}{definitions}]%
	{Definition}{Definition}{defstyle}{def}%
\newtcbtheorem[use counter from=Definition,crefname={theorem}{theorems}]%
	{Theorem}{Theorem}{theostyle}{theo}
\newtcbtheorem[use counter from=Definition]{theo}{Theorem}%
{
	theorem style=plain,
	enhanced,
	colframe=blue!50!black,
	colback=yellow!20!white,
	coltitle=red!50!black,
	fonttitle=\upshape\bfseries,
	fontupper=\itshape,
	drop fuzzy shadow=blue!50!black!50!white,
	boxrule=0.4pt}{theo}
\newtcbtheorem[use counter from=Definition]{DashedDefinition}{Definition}%
 {
 	enhanced,
 	frame empty,
 	interior empty,
 	colframe=darkspringgreen!50!white,
	coltitle=darkspringgreen!50!black,
	fonttitle=\bfseries,
	colbacktitle=darkspringgreen!15!white,
	borderline={0.5mm}{0mm}{darkspringgreen!15!white},
	borderline={0.5mm}{0mm}{darkspringgreen!50!white,dashed},
	attach boxed title to top center={yshift=-2mm},
	boxed title style={boxrule=0.4pt},
	varwidth boxed title}{theo}
\newtcblisting[auto counter,number within=section]{disexam}{
	skin=bicolor,
	colback=white!30!beaublue,
	colbacklower=white,
	colframe=black,
	before skip=\medskipamount,
	after skip=\medskipamount,
	fontlower=\footnotesize,
	listing options={style=tcblatex,texcsstyle=*\color{red!70!black}},}

\begin{document}
\begin{titlepage}
	\centering 
	\scshape 
	\vspace*{1.5\baselineskip} 

	\rule{13cm}{1.6pt}\vspace*{-\baselineskip}\vspace*{2pt} 
	\rule{13cm}{0.4pt} 
	
		\vspace{0.1\baselineskip} 
	{ {\Huge MORPHOLO C++ Library \\ User\vtick s Guide\\}
			\vspace{8mm}
	  {\large for glasses-free multi-view stereo vision\\ and streaming of live 3D video\\} }
		\vspace{0.75\baselineskip} 
	\rule{13cm}{0.4pt}\vspace*{-\baselineskip}\vspace{3.2pt} 
	\rule{13cm}{1.6pt} 
	
	\vspace{0.12\baselineskip} 
	{\large version 1.0 - Dec 2019 \\ by Enrique Canessa \& Livio Tenze\\
		\vspace*{1.2\baselineskip}
		\includegraphics[scale=1.4]{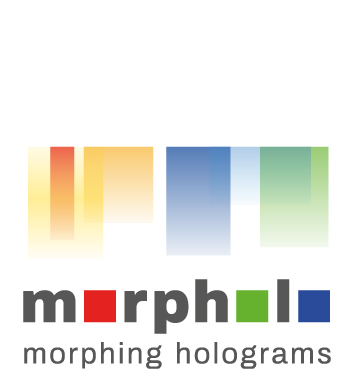}

\begin{tcolorbox}[width=0.4\textwidth, colframe=myblueii]
	 \; \url{www.morpholo.it}
\end{tcolorbox}
        }

\end{titlepage}
\tableofcontents
\vfill
\small{\noindent \textbf{About this Tutorial} \vspace{-3mm}\\
\noindent \rule{3.3cm}{0.5pt} \\
This User\vtick s Guide was written for the benefit of all teachers, students and all those interested in wanting to use the MORPHOLO C++ Libray for studies, research, demos, tests, etc. on easy and low-cost 3D Vision. The contents in this Guide are free for public use.}
\newpage
\newgeometry{
	left=29mm, 
	right=29mm, 
	top=20mm, 
	bottom=15mm}
\section{About MORPHOLO}
\vspace{3cm}

The MORPHOLO C++ extended Library allows to convert a specific stereoscopic snapshot into a Native multi-view image through morphing algorithms taking into account display calibration data for specific slanted lenticular 3D monitors. 
A slanted lenticular system allows to distribute loss of resolution in both directions, the horizontal and vertical, by slanting the structure of the lenticular lens.

\begin{center}
	\includegraphics[width=0.2\textwidth]{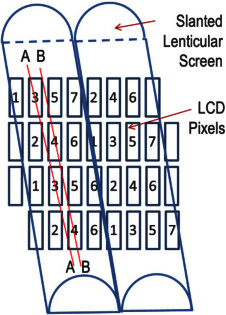} \\
    {\bf Figure 1}: {\em Slanted lenticular screen on a LCD array \\ used to enhance image quality \cite{MO1}.}
\end{center}

The MORPHOLO Library offers a fast conversion algotithm to transform stereoscopic images into reasonable holograms via the automatic implementation of a one-time configuration Lookup Table (LUT) \cite{MO0} -to replace runtime computation, and by the unsupervised morphing deformations between Left (L) and Right (R) stereoscopic images (or from a given sequential set of plain images).

\begin{center}
	\includegraphics[width=0.4\textwidth]{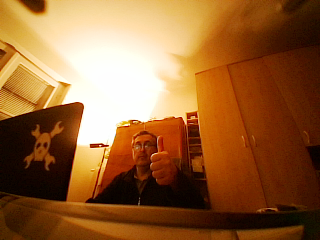}
	\includegraphics[width=0.4\textwidth]{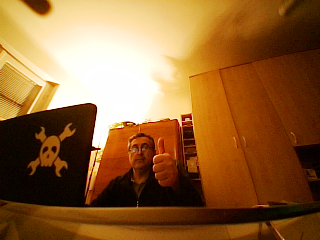} \\
    {\bf Figure 2}: {\em Typical L-R Stereoscopic images.}
\end{center}

Although morphing cannot provide complete depth information on distant regions, the generation of multi-view images starting from a pair of stereoscopic images through MORPHOLO could still offer a potential alternative method for 3D vision.
It mainly extrapolates the parallax encoded in the steroscopic images by retrieving information from nearby pixels only!

MORPHOLO can also be implemented for glasses-free live applicatons of 3D video streaming, and for diverse innovative scientific, engineering and 3D video game applications. In particular, it could be 
also useful for Panoptic studies of dynamics 3D structures, and {\em Stereomicroscopy} analysis in real-time for microsurgery or for examining microscopic samples with complex 3D topography.

\subsection{Motivation}

A 2D image is worth a thousand words. Notwithstanding a 3D reconstruction of a scene is worth a million \cite{MO1}. 

Multi-views 3D display technology is still under development even if the 3D TV technology was emerging together with the raising of the XXI century. It still needs to be optimized so as to reduce the high production costs and some of the health concerns induced by such 3D presentations. 

The new class of slanted lenticular 3D displays, such as the standalone HDMI Looking Glass HoloPlay Monitors (\url{www.lookingglassfactory.com}), allow to display a hologram of simultaneous $N\times M$ different views at $60 fps$ formed through a collage of images (or Quilt as in Fig.8) without the need of wearing unnatural headsets. These devices combine light-field and volumetric technologies, and allow to visualize the complexity of real world objects and 3D scenarios for around 40\degree to 50\degree fields of view (FOV). These 3D screens have specific computer display calibration values each for a correct rendering.

Motivated by these novel technologies, and by an earlier study on animating historical stereograms with optical flow morphing \cite{MO2}, we have developed the MORPHOLO Library to reconstruct reality from stereoscopic images. The Library is wrtten in C++ and implemented using the openCV Library with the goal of minimizing the image processing time to approach real time applications in 3D streaming using low-cost hardware (i.e., simple DIY or available stereo webcams) without the need for the viewer to wear any special 3D glasses.

Full technical description of the MORPHOLO project including classes, structs, unions and interfaces with brief descriptions, can be found at \url{www.morpholo.it} or downloading the {\tt ".deb"} package.

\begin{center}
                \includegraphics[width=0.6\textwidth]{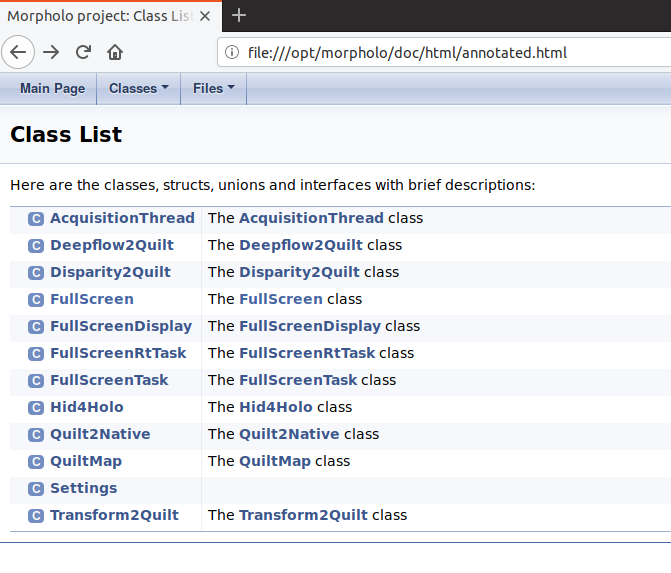} \\
    {\bf Figure 3}: {\em Class List of MORPHOLO Library.}
\end{center}

\subsection{Features}

MORPHOLO command lines lead to extend observations of 2D stereoscopic images (or, alternatively, from a given sequential set of plain images) virtual interpretations of reality --at least, within a wide FOV. 

In other words, MORPHOLO C++ Library provides the following features:
\begin{itemize}
     \item {\bf Communication with the slanted lenticular PC screen to get its calibration data.} \\
{\em (i)} To read display calibration values via a specific {\tt eeprom} interface \cite{MO3a} (i.e., read/write {\tt ".json"} file). 
 \\
{\em (ii)} To generate an associated LUT from these values and also according to some given resolution for the screen, FOV, and other variables in the {\tt ".json"} file.
     \item {\bf The transformation of a stereoscopic image to a collection of views by morphing} (that all together could form a single 3D video frame). \\
{\em (iii)} Retrieve synchronous stereoscopic images taken from a stereo webcam. \\
{\em (iv)} Generate $n$ intermediate views by morphing. \\
{\em (v)} Place these views into a single standard Quilt image.
     \item {\bf The mapping of these multi-views to the destination holographic space.} \\
{\em (vi)} Treat the collection of views in the Quilt into Native images consisting of light-fields that can be correctly displayed on the specific slanted, lenticular screen.
\end{itemize}

\subsection{Copyright}

\begin{tcolorbox}[width=1\textwidth, colframe=myblueiii]
\textcopyright Permission to use, copy, and distribute the MORPHOLO C++ Library and its documentation for educational purposes ONLY, and without any fee, is hereby granted provided that reference to \url{www.morpholo.it} appears in all distributed copies and in all supporting documentation.  This Library is provided {\em "as is"} without any express or implied warranty.
\end{tcolorbox}

\subsection{Credits and Contacts}

MORPHOLO C++ Library for Morphing Holograms is developed and maintained by Enrique Canessa and Livio Tenze from the ICTP Science Dissemination Unit (SDU), in Trieste, Italy.

For further information, get binaries, papers, presentations, manuals, etc., or to report Bugs, please contact us at: \href{mailto:sdu@ictp.it}{"sdu@ictp.it"} or visit our project website: \url{www.morpholo.it}

\newpage
\section{Theory}
\vspace{3cm}

\subsection{Stereo Image Capture}

Lenticular, autostereoscopic multi-view displays require multiple views of a scene to provide motion parallax and get a realistic 3D experience. By changing the viewing angle, different stereoscopic pairs are perceived.  However, capturing arbitrary number of views can be cumbersome, and in some occasions impossible due., e.g., to occlusion \cite{MO3b}.

Instant stereoscopic images can be taken using a set of parallel and toe-in cameras horizontally separated as depicted in the next figure \cite{MO33a}. Convergence avoids cutting off lateral parts of the two images to fulfill frame images in stereo display. 
In the parallel-cameras method, the visual camera axes are parallel, making the convergence distance $z$ to be infinite.

Stereo images captured with toe-in cameras may result in geometric distortion due to the projection difference. Vertical parallax can be introduced by converged cameras causing a change in the vertical disparities at the viewer"s retinas and are likely to affect the 3D perception \cite{MO7a}. Perspective correction and alignment are done during computer graphic rendering \cite{MO7}. 

\begin{center}
		\includegraphics[width=1\textwidth]{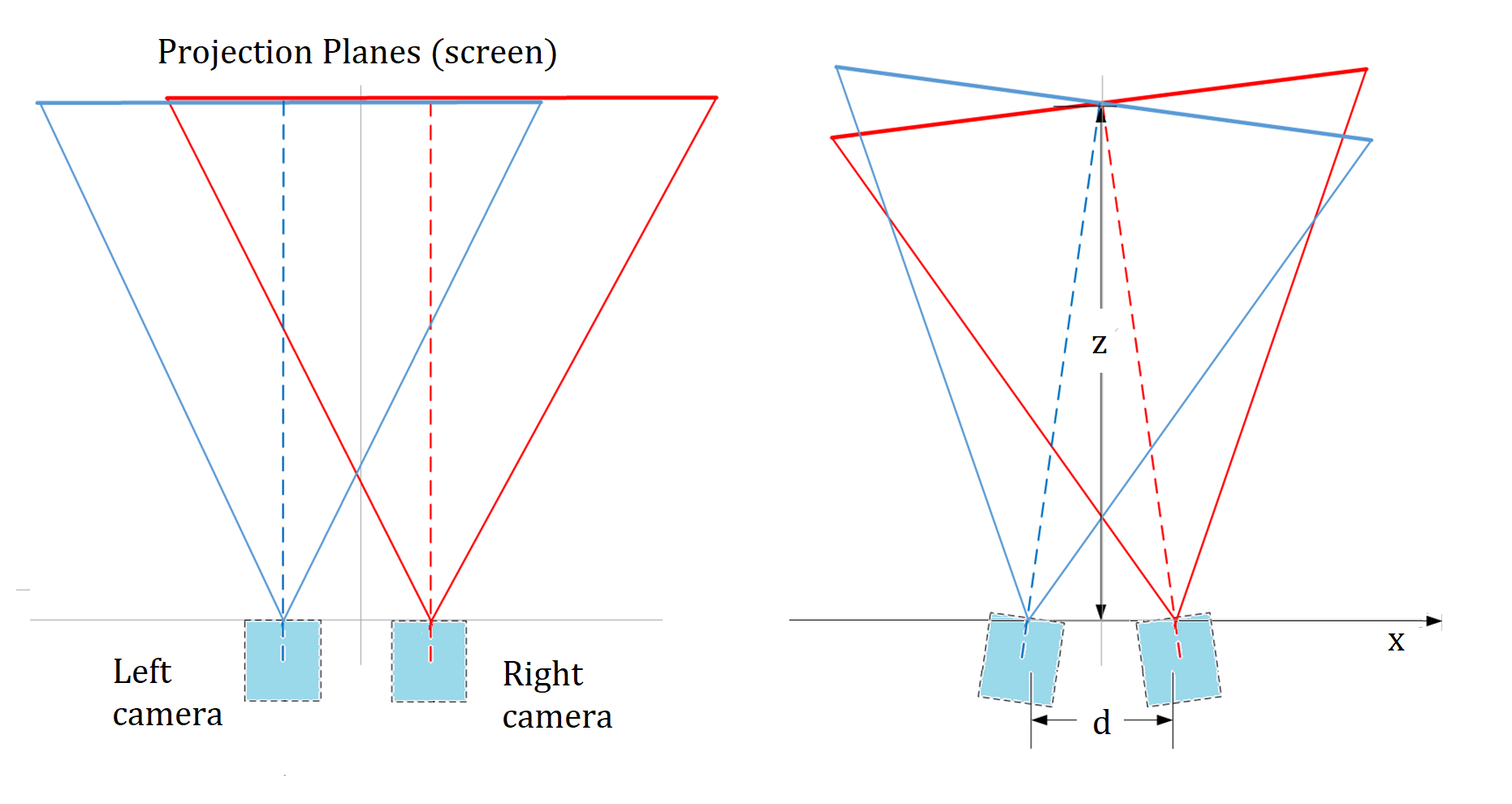}
    {\bf Figure 4}: {\em Capture configurations for stereo systems with a horizontal baseline: \\ Top view of Parallel cameras (left) and Converging cameras, also called toe-in.}
\end{center}

\newpage
Notwithstanding such constraints, we have verified via {\em trial-and-error approach} that \cite{MO8}:

\begin{tcolorbox}[width=1\textwidth, colframe=myblueiii]
using {\em slightly} toe-in cameras, they actually produce a 3D image that has more depth as compared to using paralled cameras images.  This is so when the Native 3D image produced by MORPHOLO is visualized in slanted lenticular LCD displays (such as the Looking Glass Holoplay). 
\end{tcolorbox}

In fact, slanted lenticular-based stereoscopic 3D displays provide only horizontal parallax \cite{MO1}. They generally lack vertical parallax, hence any vertical parallax introduced using stereo toe-in cameras gets reduced.

The projection convergence allows to create the illusion of a captured multi-view scene having much more 3-dimensional depth.

\subsection{Morphing Algorithms}

Within MORPHOLO, one can automatically generate a given number of intermediate views (having different viewing angles and needed to create native holographic frames from Quilts, by selecting from two different morphing techniques: (i) Disparity Map and (ii) DeepFlow. As explained below, View Morphing has not been included since it presents limits for real-time 3D streaming. Same as for tri-view morphing which follow the view morphing approach as in \cite{MO8a}.

\begin{itemize} 
\item Disparity Morphing: A method that depends on the estimated disparity map for each 
image pair based on the epipolar geometry \cite{MO9,MO9a}. \\
The epipolar constraints speed up the triangulation of the corresponding reference points 
in the L and R image. The non-linear 
forward and backward morphing functions are defined based on the estimated disparity maps
with some interpolation function pre-defined to deal with occlusion.
For smaller disparity, the darker a point is and further away it is from the camera.

\begin{center}
	\includegraphics[width=0.4\textwidth]{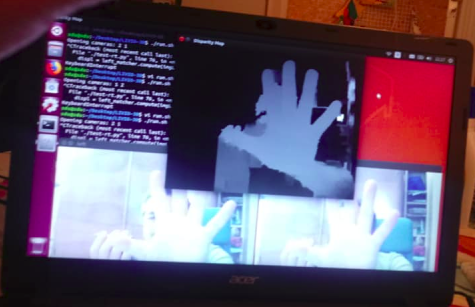} \\ 
    {\bf Figure 5}: {\em Disparity Map from stereo images with a color coding.}
\end{center}

\item DeepFlow: A variational matching approach for optical flow \cite{MO10}. \\
Optical flow computation is a key component in many computer vision systems 
designed for tasks such as action detection or activity recognition overcoming 
problems that arise in realistic videos such as: motion discontinuities, occlusions, 
illumination changes and ability to deal with (large) displacements.

\begin{center}
	\includegraphics[width=0.4\textwidth]{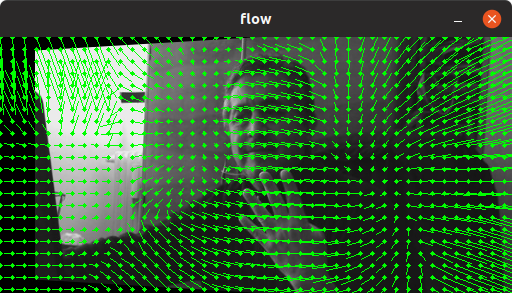} 
	\includegraphics[width=0.4\textwidth]{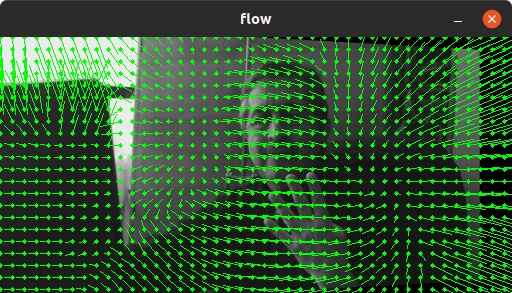} \\
    {\bf Figure 6}: {\em Optical flow, or vector field, describing movement \\ between stereoscopic images.}
\end{center}

\item View Morphing: An extension of existing image morphing tools for transitions between images by adding image prewarping and postwarping steps \cite{MO11}. \\
It can handle 3D projective camera and scene transformations,
creating more natural transitions between images at the expense of long execution time.
This algorithm is both complex and slow, making it computational 
extensive for real-time 3D video streaming. This is because of the
difficulty of encapsulating motion with certain number of keypoints or matching correspondences
(given manually or predicted by deep learning algorithms), during the dynamics of a scene.

\begin{center}
	\includegraphics[width=0.46\textwidth]{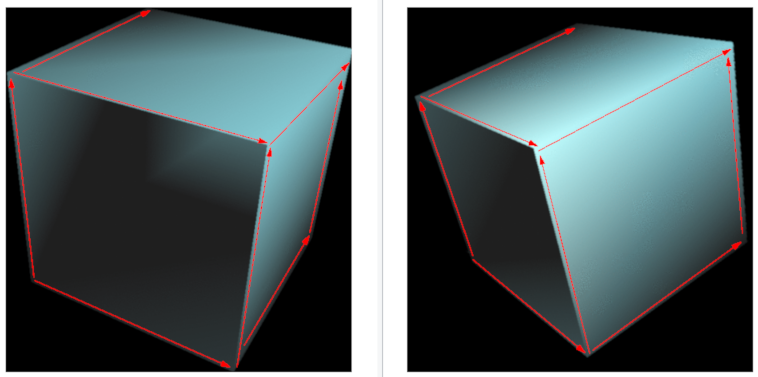} 
	\includegraphics[width=0.4\textwidth]{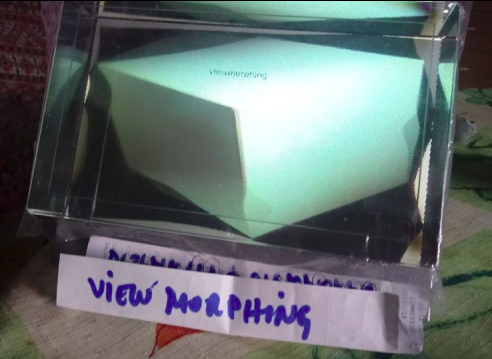} \\
    {\bf Figure 7}: {\em Example of point correspondences in steroscopic images, and MORPHOLO Native output on a Holoplay using View Morphing.}
\end{center}
\end{itemize}

\subsection{The Quilt}

The intermediate views generated using MORPHOLO can be then added into a Quilt collage sequentially as shown below. The fading out from L to R images, aims to render a whole panoramic scene into just a given number of in-between snapshots.

\begin{center}
	\includegraphics[width=0.6\textwidth]{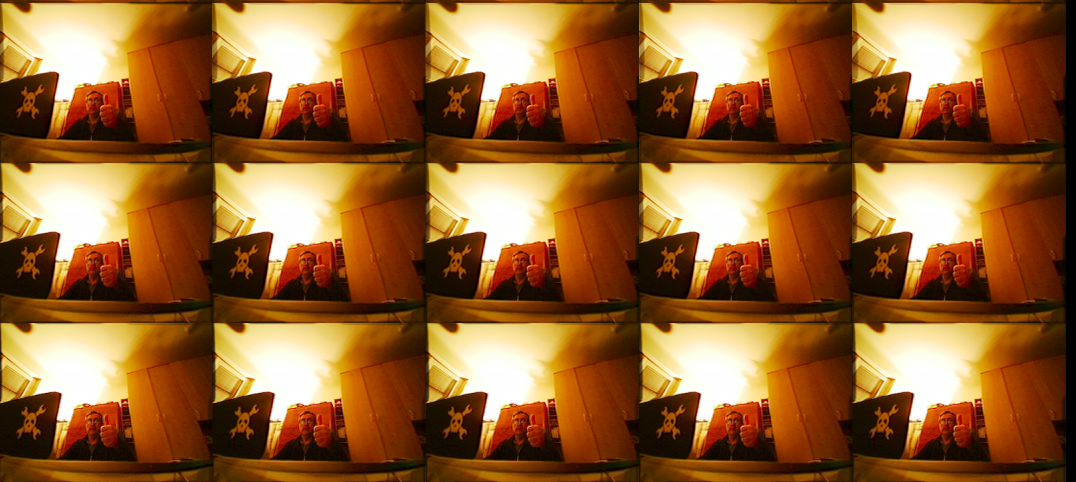} \\
    {\bf Figure 8}: {\em Central part of a 9x5 Quilt generated by morphing \\ starting from the stereoscopic images in Fig.2.}
\end{center}

The CPU time for creating such sets of images, and especially the holographic multi-view outputs (i.e., figure 4), varies considerably between diffrent morphing algorithms. Key processing steps in the DeepFlow field include the arbitrary, and time-consuming, matching by polynomial interpolation to approximate neighborhood pixel intensities, warping and optimization. Optical flow techniques are sensitive to the presence of occlusions, illumination variations and out-of-plane movements, whereas disparity map leads to obtain smoother and faster translation motion between consecutive images. 

However, according to our experiences: 

\begin{tcolorbox}[width=1\textwidth, colframe=myblueiii]
DeepFlow morphing alghorithm leads by far to a better and acceptable illusion of depth and parallax in the horizontal direction when using toe-in webcams. 
\end{tcolorbox}

The $N\times M$ morphed images in this figure --forming the Quilt, are converted into a native light-field image via Eq.(1) as shown in the figure below.

\subsection{Light-Fields and Native Images}

A generic expression for the relation between the pixels of a slanted lenticular 3D LCD and the multiple perspective views was first derived in \cite{MO3}. Each sub-pixel on the 3D-LCD is mapped to a certain view number and color value (i.e., in the lightfield domain). If $i$ and $j$ denote the panel coordinates for each sub-pixel, then

\begin{equation}
N_{i,j} = N_{tot} (i - i_{off} - 3_{j} tan(\alpha)) mod(P_{x})/P \;\;\; ,  \label{eq:1}
\end{equation}

where $N$ denotes the view number of a certain viewpoint, $\alpha$ the slanted angle between the lenticular lens and the LCD panel and $P_x$ the lenticular pitch. Upscaling multiple views (e.g., upscaling from the Quilt to the Native Looking Glass image) requires lots of CPU resources and increases system complexity \cite{MO4,MO5,MO6}. With the generation of a device-dependent LUT table this time can be reduced considerably.

\begin{center}
	\includegraphics[width=0.4\textwidth]{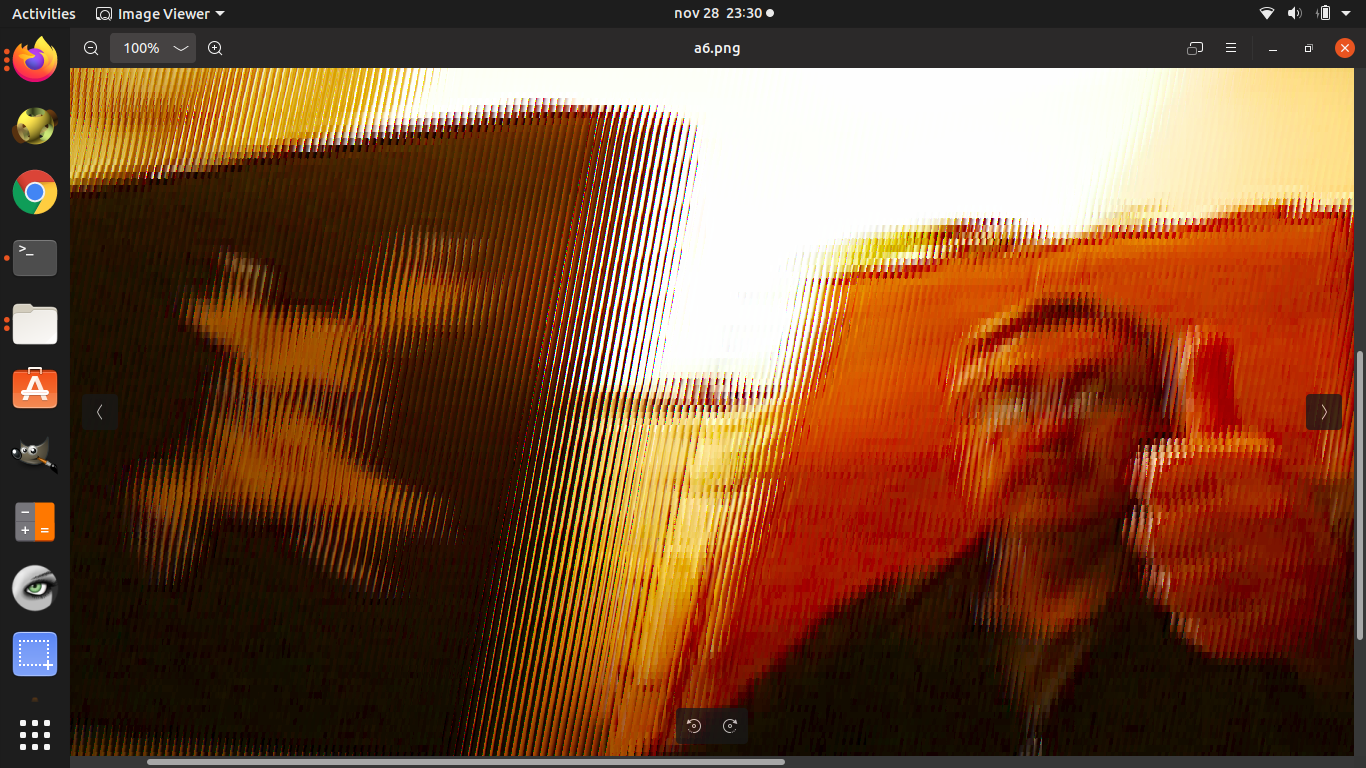} \\
    {\bf Figure 9}: {\em Example of multi-view hologram output by MORPHOLO Library \\ starting from morphed stereoscopic images. See also video demo at: \url{https://www.youtube.com/watch?v=6FAhmI-vtLQ}}
\end{center}

MORPHOLO C++ Library creates multiview images via Eq.(\ref{eq:1}) starting from a morphed stereoscopic scene (or, alternatively, from a given set of sequential photos) to form the rich Native light-field image.

\newpage
\section{Technical Notes / Requirements}
\vspace{4cm}

The simplest (DIY or available) harware needed so far to implement MORPHOLO is as follows:

\subsection{Computer with Linux O.S. and Slanted Lenticular Display}

\begin{itemize}
   \item The PC used were standard Intel Core (i3-i5, 64bit, 4-8G RAM), 1366x768p resolution, with
graphics card Nvidia GeoForce 820M output 2560x1600p and O.S. Linux Ubuntu 19.04. 

   \item The Looking Glass HoloPlay in Fig.10 is an external HDMI video monitor with slanted lenticular display. It provides a novel glasses-free way to preview 3D objects and scenes within a FOV 40\degree to 50\degree. Each Looking Glass holds own calibration ({\tt ".json"}) data for correct rendering.

\begin{center}
	\includegraphics[width=0.4\textwidth]{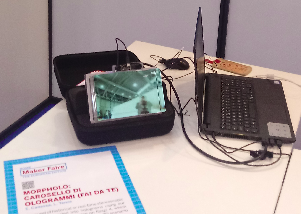} 
	\includegraphics[width=0.363\textwidth]{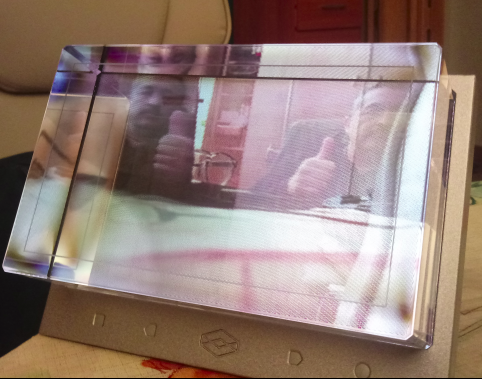} \\
    {\bf Figure 10}: {\em Headset-free Looking Glass 3D Holoplay monitor. \\ \url{www.lookingglassfactory.com}}
\end{center}
\end{itemize}

\subsection{Stereo Camera with Parallel or Convergent Lens}

\begin{itemize}
   \item Dual lens USB 3.0 camera: A compact ELP-960P2CAM stereo webcam with no distortion dual lens aligned parallel as in Fig.11.
Its two camera video frames are synchronous with 1280x960p max. resolution with a large FOV. This feature enables us to 
observe by one sigle shot a scene from two different viewpoints without the need for any prior (L-R images) calibration.

\begin{center}
	\includegraphics[width=0.34\textwidth]{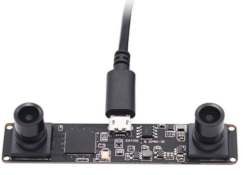} \\
    {\bf Figure 11}: {\em Compact, low-cost ELP synchronized stereo webcam. \url{www.lookingglassfactory.com}}
\end{center}

\vspace{0.5cm}
Or, alternatively,

\vspace{0.5cm}
   \item Building the stereo camera by two modular USB webcam in parallel. These need to be aligned, calibrated, etc at different separations and the video from both perspectives are retrieved in two separate windows. To get started with this see for example \cite{MO14}.

\begin{center}
        \includegraphics[width=0.3\textwidth]{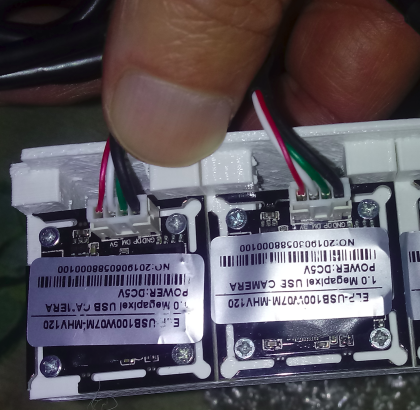} \hspace{0.04cm}
        \includegraphics[width=0.32\textwidth]{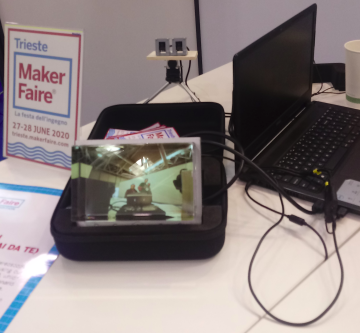} \\
    {\bf Figure 12}: {\em Pair of modular webcams in the Holoplay display.}
\end{center}

\vspace{0.5cm}
Or, alternatively,

\vspace{0.5cm}
   \item One modular USB webcam only with mirrors. Mirrors can be used to convert a single camera into a pseudo-stereo imaging system 
to view and record an object or scene from two different angles, one in each half of 
the CMOS camera sensor \cite{MO12,MO13}.
3D systems with mirror and prisms are extensively used for 3D macro photography \cite{MO15}.

\begin{center}
	\includegraphics[width=0.24\textwidth]{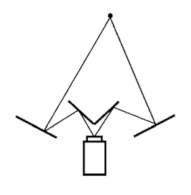} 
	\includegraphics[width=0.24\textwidth]{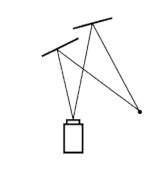} \hspace{0.5cm} 
	\includegraphics[width=0.18\textwidth]{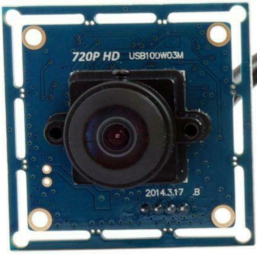} \\
    {\bf Figure 13}: {\em Single camera stereo systems using two and four planar mirrors \cite{MO13}.}
\end{center}

\vspace{0.5cm}
Or, alternatively,

\vspace{0.5cm}
   \item Two modular USB webcam convergent or toe-in (to improve the 3D) at close separation.

\begin{center}
	\includegraphics[width=0.4\textwidth]{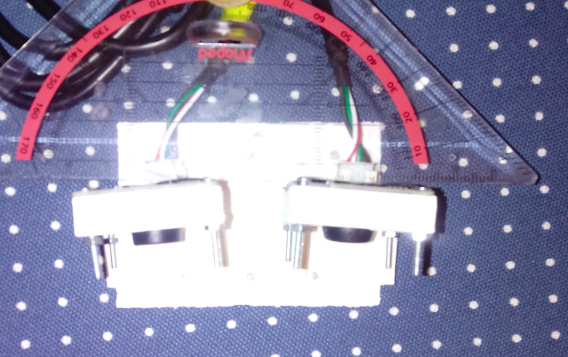} 
	\includegraphics[width=0.36\textwidth]{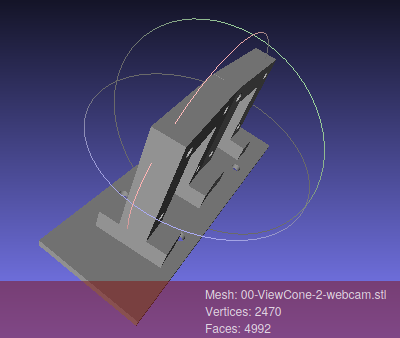} \\
    {\bf Figure 14}: {\em Convergent DIY stereo webcam with FOV of 120\degree each lens and toe-in angle of 5\degree. Graphics of the OpenSCAD {\tt ".stl"} file used for the holder (right).}
\end{center}

As mentioned before, we have verified that {\em slightly} toe-in cameras produce a 3D image that has more depth as compared to using paralled cameras images. This is so when the Native 3D image produced by MORPHOLO is visualized in slanted lenticular LCD displays since vertical parallax gets reduced \cite{MO1}.
A toe-in angle of 5\degree between the two webcam is enough for a FOV of 120\degree each with lens baseline of $4\; cm$ (about $38\%$ smaller than the human eyes separation).
\end{itemize}

\newpage
\section{How to Install {\tt morpholo-x.x.x-Linux.deb}}
\vspace{3cm}

The latest version of the Debian {\tt morpholo-x.x.x-Linux.deb} package can be downloaded from \url{www.morpholo.it} A connection to the Internet is needed to download new releases. 

\subsection{Dependencies}

It is necessary to install first some extra packages and their dependencies. The following packages (and their dependencies) are needed to be installed before hand: 

\begin{tcolorbox}[width=1\textwidth, colframe=lightgray]
libopencv-videoio3.2\, libopencv-imgproc3.2\, libopencv-core3.2\, libopencv-imgcodecs3.2\, libopencv-contrib3.2\, libopencv-calib3d3.2\, libopencv-imgcodecs3.2\, libopencv-video3.2\, libopencv-flann3.2\, libqt5widgets5\, libqt5gui5\, libqt5core5a\, libhidapi-libusb0
\end{tcolorbox}

These could become out of date, so check the list of needed packages by issuing the following command:

\begin{tcolorbox}[width=1\textwidth, colframe=myblueii]
{\tt 
    dpkg -I morpholo-x.x.x-Linux.deb
}
\end{tcolorbox}

In {\tt "Depends: $\cdots$"} you can find the updated list of required packages.

To install the required packages (listed above) issue the command: {\tt "sudo apt-get install <pkg1> <pkg2> $\cdots$"} and so on. For example, 

\begin{tcolorbox}[width=1\textwidth, colframe=myblueii]
{\tt 
 sudo apt-get install libopencv-videoio3.2\, libopencv-imgproc3.2\, $\cdots$
}
\end{tcolorbox}

\subsection{Library Install}

To install the MORPHOLO ({\tt ".deb"}) package in the {\tt /opt/morpholo} directory type the command line: 

\begin{tcolorbox}[width=1\textwidth, colframe=red]
\textcolor{red}{{\tt 
   sudo dpkg -i morpholo-x.x.x-Linux.deb
}}
\end{tcolorbox}

In case some {\tt "warning: files list file for package $\cdots$"} may appear on the screen after having issued {\tt "dpkg"}, then type 

\begin{tcolorbox}[width=1\textwidth, colframe=myblueii]
{\tt 
sudo apt-get install -f
}
\end{tcolorbox}

which install what is missing to {\tt "dpkg"}.

{\bf It also necessary to set the path to the working directory before the MORPHOLO commands can be used}. Then, type:

\begin{tcolorbox}[width=1\textwidth, colframe=red]
\textcolor{red}{{\tt 
sudo -i \\
cd /opt/morpholo/bin/ \\
source /opt/morpholo/bin/setupvars.sh 
}}
\end{tcolorbox}

which defines

\begin{tcolorbox}[width=1\textwidth, colframe=lightgray]
INSTALLDIR="/opt/morpholo" \\
export PATH=\$INSTALLDIR/bin:\$\{PATH\} \\
export LD\_LIBRARY\_PATH=\$INSTALLDIR/lib:\$\{LD\_LIBRARY\_PATH\}
\end{tcolorbox}

\subsection{Uninstall}

In order to remove the MORPHOLO ({\tt ".deb"}) package type 

\begin{tcolorbox}[width=1\textwidth, colframe=myblueii]
{\tt 
sudo dpkg -r morpholo 
}
\end{tcolorbox}

Then check that the {\tt /opt/morpholo} directory becomes empty! 

\newpage
\section{Using MORPHOLO}
\vspace{3cm}

\subsection{Extract Calibration Data from 3D Display}

Each Looking Glass HoloPlay (shown in Fig.10) holds its own per-device calibration data 
set in the phase of manufacturing for correct rendering!
To get this data (in a {\tt ".json"}-type of text file), connect the HoloPlay device to your PC and
issue the command as root:

\begin{tcolorbox}[width=1\textwidth, colframe=myblueii]
{\tt 
sudo ./morph2native -m DEFAULT\_MAP\_FILE -c CALIBRATION\_FILE   
\newline
\newline
with 
\begin{addmargin}[2em]{1em}
\begin{tabular}{lll}
 -h,   &   --help                            &   Displays help     \\
 -m,   &   --map <Map file>                  &   Save map to file  \\
 -c,   &   --calibration <Calibration file>  & Save calibration to file 
\end{tabular}
\end{addmargin}
}
\end{tcolorbox}

This command generates two files: (i) one with a default LUT mapping (that can later be also re-estimated), and (ii) another with the unique calibration data for the particular HoloPlay being used. 

For example, the above command can look like this:

\begin{tcolorbox}[width=1\textwidth, colframe=myblueiii]
./morph2native -m default\_map.map  mycal.json
\end{tcolorbox}

The {\tt CALIBRATION\_FILE} (i.e., {\tt "mycal.json"} in the example) will look like this:

\begin{tcolorbox}[width=1\textwidth, colframe=myblueiii]
{"configVersion":"1.0","serial":"LKG-2K-02491","pitch":{"value":47.56159591674805},
"slope":{"value":-5.5113043785095219},"center":{"value":-0.09782609343528748},
"viewCone":{"value":40.0},"invView":{"value":1.0},"verticalAngle":{"value":0.0},"DPI":{"value":
338.0},"screenW":{"value":2560.0},"screenH":{"value":1600.0},"flipImageX":{"value":0.0},
"flipImageY":{"value":0.0},"flipSubp":{"value":0.0}}
\end{tcolorbox}

The binary {\tt DEFAULT\_MAP\_FILE} file contains the LUT mapping for a default 8x4 Quilt whose single image is assumed to be size $256\times512 px$. As shown next this {\tt ".map"} file can be modified for different Quilt sizes, adapted to the {\tt CALIBRATION\_FILE} file.

\subsection{Create LUT Mapping from Calibration File and Quilt Inputs}

Use:

\begin{tcolorbox}[width=1\textwidth, colframe=myblueii]
{\tt 
\begin{tabular}{rl}
./morph2native & -r RESOLUTION -q NxM\_QUILT  \\ 
     & \hspace{2cm} -m FINAL\_MAP\_FILE  CALIBRATION\_FILE 
\end{tabular}
\newline
\newline
with 
\begin{addmargin}[2em]{1em}
\begin{tabular}{lll}
 -h,   &   --help                         &   Displays help     \\
 -r,   &   --resolution <rows x columns>  &   Image resolution for quilt   \\
 -q,   &   --quilt <rows x columns>       &   Mask for quilt   \\
 -m,   &   --map <Map file>               &   Save map to file 
\end{tabular}
\end{addmargin}
}
\end{tcolorbox}

For example, the above command can look like this:

\begin{tcolorbox}[width=1\textwidth, colframe=myblueiii]
./morph2native -r 240x320 -q 9x5 -m mymap.map  mycal.json
\end{tcolorbox}

\subsection{Generate a Quilt}

Up scaling multiple views from the stereo L-R images to the Native image can require substantial computer resources depending on the Quilt resolution. 
Each point of these Quilts has to be mapped to the destination Native holographic image of 2560x1600p. The mapping procedure needs to be very accurate according to the Quilt dimensions and has to take into account the specific display calibration data. Within MORPHOLO, it is possible to get the Quilt from 
different sources and by selecting from two different morphing techniques: (i) Disparity Map and (ii) DeepFlow. 

\subsubsection{Quilt from stereo L-R images}

{\tt morph2quilt} is a Disparity map- and DeepFlow-based tool to generate intermediate views and the Quilt, using archived left and right stereoscopic images.

\begin{tcolorbox}[width=1\textwidth, colframe=myblueii]
{\tt 
./morph2quilt -l IMG1 -r IMG2 -m NxM\_QUILT -d -s <factor> QUILT\_FILE
\newline
\newline
with 
\begin{addmargin}[2em]{1em}
\begin{tabular}{lll}
  -h, &  --help                              &  Displays help          \\
  -l, &  --left <Left image path>            &  Input left image       \\
  -r, &  --right <Right image path>          &  Input right image      \\
  -m, &  --mask <rows x columns>             &  Mask for quilt         \\ 
  -d, &  --deep                              &  Deepflow mode          \\
  -s, &  --subsampling <Subsampling factor>  &  Speed up morphing
\end{tabular}
\end{addmargin}
}
\end{tcolorbox}

By default, Disparity map is used for the morphing, 
otherwise use the option {\tt "-d}" to apply the DeepFlow algorthim.
The option {\tt "-s"} sets a subsampling in the morphing process to speed up 
calculations, e.g., use small values {\tt "-s 2"}. 

For example, to create a Quilt by DeepFlow the above command can look like this:

\begin{tcolorbox}[width=1\textwidth, colframe=myblueiii]
./morph2quilt -l photo2.png -r photo4.png -m 9x5 -d myquilt\_stereo\_imgs.png
\end{tcolorbox}

\subsubsection{Quilt from single stereo camera with parallel lens}

A stereo camera with no distortion dual lens, aligned parallel as in Fig.11,
together with a single modular USB webcam with mirrors as in Fig.13,
can be used as a pseudo-stereo imaging system.  In this case, you can
use the standard applications {\tt ffmpeg} and {\tt convert} to capture images.
For example, use:

\begin{tcolorbox}[width=1\textwidth, colframe=myblueiii]
{\tt 
\begin{tabular}{rl}
	ffmpeg  &  -f v4l2 -framerate 10 -thread\_queue\_size 512 -i /dev/video2  \\
                & \hspace{0.8cm}  -vf "rotate=0" -ss 00:00:01 -vframes 1 -map 0 stereo.png  \\
                &        \\
	convert &  -crop 50\%x100\% stereo.png camera.png   \\
 	mv   &   camera-0.png camera2.png   \\
	mv   &   camera-1.png camera4.png    \\
        convert   &   -resize 320x240 camera2.png camera2.png \\
        convert   &   -resize 320x240 camera4.png camera4.png 
\end{tabular}
}
\end{tcolorbox}

Then, as above, issue for example the command

\begin{tcolorbox}[width=1\textwidth, colframe=myblueiii]
./morph2quilt -l camera2.png -r camera4.png -m 9x5 -d myquilt\_single\_stereocam.png
\end{tcolorbox}

\subsubsection{Quilt from two different parallel or convergent (toe-in) camera}

In this case, type this long command

\begin{tcolorbox}[width=1\textwidth, colframe=myblueiii]
{\tt 
\begin{tabular}{rl}
ffmpeg &  -f v4l2 -framerate 10 -video\_size RESOLUTION  \\
       & \hspace{0.2cm} -thread\_queue\_size 512 -i /dev/video2 -vf "rotate=0" \\
       & \hspace{0.6cm} -ss 00:00:01 -vframes 1 -map 0 camera4.png \\
       &  -f v4l2 -framerate 10 -video\_size RESOLUTION  \\
       & \hspace{0.2cm} -thread\_queue\_size 512 -i /dev/video4 -vf "rotate=0" \\
       & \hspace{0.6cm} -ss 00:00:01 -vframes 1 -map 1 camera2.png 
\end{tabular}
}
\end{tcolorbox}

with, e.g., input {\tt RESOLUTION} equal to $240\times 320$.  Then, as above, issue the command
 
\begin{tcolorbox}[width=1\textwidth, colframe=myblueiii]
./morph2quilt -l camera2.png -r camera4.png -m 9x5 -d myquilt\_two\_cam.png
\end{tcolorbox}

\subsection{Generate Native Image / Multi-Vision Stereo Vision}

The MORPHOLO Library commands, that allows to convert the Quilt into a Native multi-view image --through morphing algorithms and taking into account display calibration data for specific slanted lenticular 3D monitors, are the following.

\subsubsection{Native image from sorted views stored in a directory}

Use:

\begin{tcolorbox}[width=1\textwidth, colframe=myblueii]
{\tt
./images2native -q NxM\_QUILT -m FINAL\_MAP\_FILE DIR\_PATH NATIVE\_FILE
\newline
\newline
with
\begin{addmargin}[2em]{1em}
\begin{tabular}{lll}
 -h,   &   --help                         &   Displays help     \\
 -q,   &   --quilt <rows x columns>       &   Mask for quilt   \\
 -m,   &   --map <mapfile>                &   Path to the map file 
\end{tabular}
\end{addmargin}
}
\end{tcolorbox}

where {\tt DIR\_PATH} is the Path to the directory where $N\times M$ sorted 
views (or pre-loaded images) can be found. 

For example, the above command can look like this:

\begin{tcolorbox}[width=1\textwidth, colframe=myblueiii]
\begin{tabular}{rl}
./images2native  &  -q 9x5 -m mymap.map \\
     & \hspace{2cm} /home/mymorpholo \, mynative\_nxm\_imgs.png 
\end{tabular}
\end{tcolorbox}

\subsubsection{Display Native image from (historical) stereo L-R images}

The following command creates a Native image starting from a given set of 
L-R images and shows it in an external monitor via multi screen configuration:

\begin{tcolorbox}[width=1\textwidth, colframe=myblueii]
{\tt
\begin{tabular}{rl}
./morpholo-display &  -l IMG1 -r IMG2 -m NxM\_QUILT -q NxM\_QUILT \\
              &  \hspace{2cm} -d -s <factor> --screen 1
\end{tabular}
\newline
\newline
with
\begin{addmargin}[2em]{1em}
\begin{tabular}{lll}
  -h,  & --help                            &  Displays help          \\
  -l,  & --left <Left image path>          &  Input left image        \\
  -r,  & --right <Right image path>        &  Input right image      \\
  -m,  & --map <File path>                 &  Map quilt to native   \\
  -q,  & --quilt <rows x columns>          &  Mask for quilt  \\
  -d,  & --deep                            &  Deepflow mode             \\
  -s,  & --subsampling <Subsampling factor> &  Speed up morphing  \\
  --screen  &  <Integer number display>      &  Screen Nr. (0 or 1)
\end{tabular}
\end{addmargin}
}
\end{tcolorbox}

By default, {\tt "-screen 0"} and disparity map is used for the morphing,
otherwise use the option {\tt "-d"} to apply the DeepFlow algorthim.
The option {\tt "-s"} sets a subsampling in the morphing process to speed up
calculations, e.g., use small values {\tt "-s 2"}.

For example, the above command can look like this (press {\tt "Q"} to quit):

\begin{tcolorbox}[width=1\textwidth, colframe=myblueiii]
\begin{tabular}{rl}
./morpholo-display & -l photo2.png -r photo4.png -m mymap.map -q 9x5 \\
       &   \hspace{4cm} -d -s 2 -\,-screen 1
\end{tabular}
\end{tcolorbox}

\subsubsection{Native image output from input Quilt}

To obtain the input Native image for its 3D vision in slanted lenticular displays use:

\begin{tcolorbox}[width=1\textwidth, colframe=myblueii]
{\tt
\begin{tabular}{rl}
./morph2native &  -q NxM\_QUILT -r RESOLUTION -a FINAL\_MAP\_FILE  \\
     & \hspace{1.4cm} INPUT\_QUILT\_FILE \, NATIVE\_FILE
\end{tabular}
\newline
\newline
with
\begin{addmargin}[2em]{1em}
\begin{tabular}{lll}
 -h,   &   --help                         &   Displays help     \\
 -q,   &   --quilt <rows x columns>       &   Mask for quilt   \\
 -r,   &   --resolution <rows x columns>  &   Image resolution for quilt   \\
 -a,   &   --apply <Map filename>         &   Apply map to given image   
\end{tabular}
\end{addmargin}
}
\end{tcolorbox}

For example, the above command can look like this:

\begin{tcolorbox}[width=1\textwidth, colframe=myblueiii]
./morph2native -q 9x5 -r 240x320 -a mymap.map myquilt\_xyz.png mynative.png 
\end{tcolorbox}

\subsubsection{Native image display for continuos Loop presentations}

The MORPHOLO command to show a native image when multiple displays are present is: 

\begin{tcolorbox}[width=1\textwidth, colframe=myblueii]
{\tt
./native2display NATIVE\_FILE --screen 0 
\newline
\newline
with
\begin{addmargin}[2em]{1em}
\begin{tabular}{lll}
  -h, &  --help                              &  Displays help          \\
  --screen  &  <Integer number display>      &  Screen Nr. (0 or 1)
\end{tabular}
\end{addmargin}
}
\end{tcolorbox}

In order to display in a Loop randomly Native images (every $12 secs$ each and sequentially numbered
as $1, \cdots 14$ in the directory {\tt img}), the shell script below can be used:

\begin{tcolorbox}[width=1\textwidth, colframe=myblueiii]
{\tt
\begin{addmargin}[2em]{1em}
\#!/bin/bash \\
  \\
declare secs=12 \\
declare NrImg=14 \\
declare RMD="img/2.png" \\
  \\
while [ 1 ]; do \\
  \\
for j in \$(seq 1 1 \$NrImg) \\
do \\ 
  \\
echo "Display Random Native" \\
echo "\$j \$RMD" \\
  \\
./native2display \$RMD \& \\
pid=\$!  \\
  \\
for RND in \$(shuf -i 1-\$NrImg -n 1)  \\
do  \\
RMD=img/\$RND.png  \\
done   \\
echo "Changed Native (\$j) to \$RMD"  \\
  \\
sleep \$secs  \\
kill SIGUSR1 \$pid  \\
done   \\
  \\
done  \\
\end{addmargin}
}
\end{tcolorbox}

\newpage
\section{Streaming of 3D Videos}
\vspace{3cm}

\subsection{pre-Recorded Stereoscopic Video Streaming}

3D stereoscopic videos are available, e.g., in YouTube and can be freely downloaded.
The processing time to convert such stereoscopic video frames into Native images in real time,
can be large depending on the images size and resolution. However, it can be useful 
to carry out this exercise through MORPHOLO just for testing purposes.

First, prepare the archived stereoscopic video. Use the following command to 
rescale the video with fixed width and height (e.g., $320\times 180 px$):

\begin{tcolorbox}[width=1\textwidth, colframe=myblueiii]
ffmpeg -i video\_youtube.mp4 -vf scale="320:180" video\_rescaled.mp4
\end{tcolorbox}

To retain the aspect ratio based on the width, just give height as {\tt scale="320:-1"}.

Second, create LUT map from your HoloPlay calibration ({\tt ".json"}) 
file considering a small Quilt size:

\begin{tcolorbox}[width=1\textwidth, colframe=myblueiii]
./morph2native -r 128x256 -q 4x2 -m myyoutube.map mycal.json
\end{tcolorbox}

Then, to stream the pre-recorded video issue the MORPHLO command

\begin{tcolorbox}[width=1\textwidth, colframe=myblueii]
{\tt
./morpholo-rt -c CONFIG\_INI -m myyoutube.map -q 4x2 -d -s 2 --screen 0 
\newline
\newline
with 
\begin{addmargin}[2em]{1em}
\begin{tabular}{lll}
  -h, &  --help                              &  Displays help          \\
  -c, &  --config <Path to .ini file>        &  Config file path            \\
  -m, &  --map <File path>                   &  Map quilt to native    \\
  -q, &  --quilt <rows x columns>            &  Mask for quilt              \\
  -d, &  --deep                              &  Deepflow mode               \\
  -s, &  --subsampling <Subsampling factor>  &  Speed up morphing     \\
  --screen  &  <Integer number display>      &  Screen Nr. (0 or 1)
\end{tabular}
\end{addmargin}
}
\end{tcolorbox}

By default, Disparity map is used for the morphing,
otherwise use the option {\tt "-d"} to apply the DeepFlow algorthim.
The option {\tt -"s"} sets a subsampling in the morphing process to speed up
calculations, e.g., use small values {\tt -"s 2"}.

The text {\tt CONFIG\_INI} file (.e.g., {\tt "config\_youtube.ini"}) should contain:

\begin{tcolorbox}[width=1\textwidth, colframe=myblueiii]
{\tt
\begin{addmargin}[2em]{1em}
\lbrack camera\rbrack                \\
devNumber=-1            \\
width=320               \\
height=180              \\
fps=8                   \\
file="video\_rescaled.mp4"    \\
\lbrack processing\rbrack            \\
width=256               \\
height=128              \\
\lbrack native\rbrack                \\
width=2560              \\
height=1600             
\end{addmargin}
}
\end{tcolorbox}

\subsection{Elapsed Processing Time}

Existing 3D TV video technologies are mainly based on the use of stereoscopic cameras 
to save time in the process.  By using the {\tt "mycal.json"} calibration file, and applying 
the light-field geometric transformations of Eq.(1) only, one in principle can also get 
the multi-view Native image. 
This simpler procedure requires considerable calculation power, since the final native image 
for the HoloPlay must have a resolution of $2560\times 1600 px$ with RGB color 
channels -–such that, once the pixel to be mapped is fixed, the map value for each color 
channel implies separated calculations. In essence this procedure as such makes real-time video 
in 3D difficult to achieve.

However, since this mapping matrix depends on the geometric position of each pixel, and 
on the calibration parameters of the lenticular display, the constructed Lookup Tables (LUT) by
MORPHOLO allows to reduce further the runtime computation and save processing time. As 
discussed above, this LUT is created only once at the beginning of the mapping process:
Quilt $\rightarrow$ HoloPlay image, and then used for all the Native frames to be visualized. 

In Fig.15, an estimate of the total processing time employed for the generation of MORPHOLO
holograms starting from a set of small-resolution stereoscopic images is illustrated.
As quantified in these curves, total processing time steps below 0.1 sec, in between 10-{\em to}-35 different views, can be obtained via MORPHOLO with the use of LUT. This implies that in 1 second, we can generate ten or more different Native frames, opening a gateway to achieve 3D real-time video streaming. 

\begin{center}
        \includegraphics[width=0.8\textwidth]{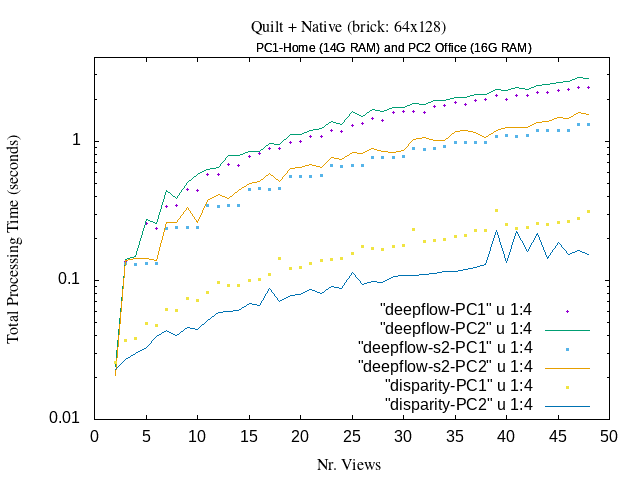} \\
    {\bf Figure 15}: {\em MORPHOLO performance test on two different PC \\ Intel Core i3-i5 with 16G-8G RAM, respectively.
}
\end{center}

The values quantified in Fig.15 can be otaining using one of this MORPHOLO commands:
{\tt morph2quilt, morph2native} and {\tt morpholo-display} by adding the option:

\begin{tcolorbox}[width=1\textwidth, colframe=myblueiii]
{\tt
\begin{addmargin}[2em]{1em}
\begin{tabular}{lll}
  -t,  & --time                            &  Show elapsed processing time    
\end{tabular}
\end{addmargin}
}
\end{tcolorbox}

For example, a shell script for estimating Quilt elapsed time in a Loop 
for variable number of views can be written as:

\begin{tcolorbox}[width=1\textwidth, colframe=myblueiii]
{\tt
\begin{addmargin}[2em]{1em}
\#!/bin/bash   \\
\#\# RUN: ./run-thiscript.sh \, 2> \, stat-morpholo-deepflow.txt   \\
  \\ 
for i in \$(seq 2 1 48)   \\
do   \\
echo "Start 1 x \$i ..."  \\
  \\ 
\#\# Quilt Elapsed Time x Deepflow:   \\
  \\ 
./morph2quilt -l photo2.png -r photo4.png -m 1x\$i -d -t myquilt\_stat.png  \\
  \\ 
done  
\end{addmargin}
}
\end{tcolorbox}

The output will be like this:

\begin{tcolorbox}[width=1\textwidth, colframe=myblueiii]
\begin{addmargin}[3em]{1em}
Start 1 x 44 ...    \\
Reading files    \\
Elapsed time (cpu time): 0.000779 s   \\
Elapsed time (wall clock): 0.000781 s   \\
Deepflow mode enabled   \\
Processing step   \\
Elapsed time (cpu time): 7.657460 s   \\
Elapsed time (wall clock): 1.982400 s  \\
Writing file  \\
Elapsed time (cpu time): 0.018364 s  \\
Elapsed time (wall clock): 0.018378 s  \\
Start 1 x 45 ... 
\end{addmargin}
\end{tcolorbox}

\subsection{3D Real-Time Video Streaming}

Create first the LUT map from your HoloPlay calibration ({\tt ".json"}) 
file considering a small Quilt size:

\begin{tcolorbox}[width=1\textwidth, colframe=myblueiii]
./morph2native -r 128x256 -q 4x2 -m mymap.map mycal.json
\end{tcolorbox}

To stream live 3D video issue the following command with one of 
the {\tt CONFIG\_INI} text file below depending on the setup choosen:

\begin{tcolorbox}[width=1\textwidth, colframe=myblueii]
{\tt
./morpholo-rt -c CONFIG\_INI -m mymap.map -q 4x2 -d -s 2 --screen 0
\newline
\newline
with
\begin{addmargin}[2em]{1em}
\begin{tabular}{lll}
  -h, &  --help                              &  Displays help          \\
  -c, &  --config <Path to .ini file>        &  Config file path            \\
  -m, &  --map <File path>                   &  Map quilt to native    \\
  -q, &  --quilt <rows x columns>            &  Mask for quilt              \\
  -d, &  --deep                              &  Deepflow mode               \\
  -s, &  --subsampling <Subsampling factor>  &  Speed up morphing     \\
  --screen  &  <Integer number display>      &  Screen Nr. (0 or 1)
\end{tabular}
\end{addmargin}
}
\end{tcolorbox}

By default, Disparity map is used for the morphing,
otherwise use the option {\tt "-d"} to apply the DeepFlow algorthim.
The option {\tt "-s"} sets a subsampling in the morphing process to speed up
calculations, e.g., use small values {\tt "-s 2"}.

\subsubsection{Using single stereo camera with parallel lens}

The {\tt CONFIG\_INI} file (.e.g., {\tt "config\_singlecam.ini"}) must contain:

\begin{tcolorbox}[width=1\textwidth, colframe=myblueiii]
{\tt
\begin{addmargin}[2em]{1em}
\lbrack camera\rbrack                \\
devNumber=2    \\
width=320               \\
height=180              \\
fps=8                   \\
\lbrack processing\rbrack            \\
width=256               \\
height=128              \\
\lbrack native\rbrack                \\
width=2560              \\
height=1600
\end{addmargin}
}
\end{tcolorbox}

\subsubsection{Using two different parallel or convergent (toe-in) camera}

In this case, the {\tt CONFIG\_INI} file (.e.g., {\tt "config\_twocam.ini"}) must contain:

\begin{tcolorbox}[width=1\textwidth, colframe=myblueiii]
{\tt
\begin{addmargin}[2em]{1em}
\lbrack camera\rbrack                \\
width=320               \\
height=180              \\
fps=8                   \\
\lbrack camera0\rbrack                \\
devNumber=2    \\
\lbrack camera1\rbrack                \\
devNumber=4    \\
\lbrack processing\rbrack            \\
width=256               \\
height=128              \\
\lbrack native\rbrack                \\
width=2560              \\
height=1600
\end{addmargin}
}
\end{tcolorbox}

\newpage
	
\addtocounter{section}{1}
\addcontentsline{toc}{section}{\protect\numberline{\thesection}~~~ References}
\end{document}